\documentclass{article}
\usepackage{booktabs}
\usepackage{graphicx}
\usepackage{amsmath}
\usepackage{amsfonts}
\usepackage[colorlinks=true, allcolors=blue]{hyperref}
\usepackage{algorithm}
\usepackage{algorithmic}
\usepackage{natbib}
\usepackage[capitalize]{cleveref}
\usepackage{siunitx}
\usepackage{subcaption}
\usepackage{multirow}

\usepackage{booktabs}
\usepackage{graphicx}

\usepackage[english]{babel}

\usepackage[letterpaper,top=2cm,bottom=2cm,left=3cm,right=3cm,marginparwidth=1.75cm]{geometry}

\title{Learning Globally Optimized Language Structure via Adversarial Training}
\author{Xuwang Yin \\{\texttt{xy4cm@virginia.edu}}}

\date{\today}

\begin{document}

\maketitle

\begin{abstract}

  Recent work has explored integrating autoregressive language models with energy-based models (EBMs) to enhance text generation capabilities. However, learning effective EBMs for text is challenged by the discrete nature of language. This work proposes an adversarial training strategy to address limitations in prior efforts. Specifically, an iterative adversarial attack algorithm is presented to generate negative samples for training the EBM by perturbing text from the autoregressive model. This aims to enable the EBM to suppress spurious modes outside the support of the data distribution. Experiments on an arithmetic sequence generation task demonstrate that the proposed adversarial training approach can substantially enhance the quality of generated sequences compared to prior methods. The results highlight the promise of adversarial techniques to improve discrete EBM training. Key contributions include: (1) an adversarial attack strategy tailored to text to generate negative samples, circumventing MCMC limitations; (2) an adversarial training algorithm for EBMs leveraging these attacks; (3) empirical validation of performance improvements on a sequence generation task.

\end{abstract}


\section{Introduction}

The recent advancement in natural language generation (NLG) has been predominantly driven by autoregressive models, particularly Locally Normalized Language Models (LNLMs). These models, including variations of Recurrent Neural Networks (RNNs) and the Transformer architecture, have shown remarkable capabilities in generating coherent and contextually relevant text. However, they often suffer from limitations inherent to their local scope --- optimizing the likelihood of each token solely based on its preceding context. This local perspective can result in sequences that, although locally plausible, are globally inconsistent or suboptimal, leading to limitations in capturing long-range dependencies and semantic coherence.

To address these shortcomings, researchers have explored the integration of LNLMs with Energy-Based Models (EBMs), which offer a more global view of sequence generation by leveraging an energy function parameterized across the entire sequence~\citep{wang2018learning,parshakova2019global,deng2019residual}. 

However, the application of EBMs to text-based tasks is challenged by the discrete nature of text, which hinders the use of gradient-based Markov Chain Monte Carlo (MCMC) methods for learning EBMs.
To sidestep this issue, the above work proposes a joint model $P_\theta(x) \propto P_{L M}(x) \exp \left(-E_\theta(x)\right)$ with which the energy function  $E_\theta(x)$ can be learned using \emph{noise contrastive estimation (NCE)} by training a binary classifier to discriminate text generated by $P_{L M}$ and text samples from the target data distribution. The quality of generated text from $P_{L M}$ can then be improved by resampling according to the learned energy function.

While theoretically \( P_\theta(x) \) can be trained to approximate the target data distribution \( P_\text{data} \), given that \( \text{Supp}(P_{LM}) = \text{Supp}(P_\text{data}) \), the practical effectiveness of this approach is constrained by \( P_{LM} \)'s ability to accurately capture the support of \( P_\text{data} \).
The study by \cite{bakhtin2019real} reveals that it is easy to find text of low probability under $P_{L M}$ but  high probability under  $E_\theta(x)$ by perturbing the text generated by $P_{L M}$. This observation suggests that the learned energy model has a multitude of \emph{spurious modes} in the residual space of \(P_{LM}\) (although their impact on the aforementioned resampling strategy may be limited, given that \(P_{LM}\) does not operate in this residual space).

In this work, we propose an \emph{adversarial training strategy} to address spurious modes and assess its efficacy in enhancing the sequence generation abilities of \( P_{LM} \). Unlike the aforementioned approach that trains a binary classifier to distinguish between samples from the data distribution \( P_\text{data} \) and generated text from \( P_{LM} \), our strategy involves training the energy model to contrast samples of \( P_\text{data} \) with \emph{adversarially perturbed} versions of \( P_{LM} \) samples. These perturbed samples are generated through gradient-free adversarial attacks targeted at the energy model within the discrete input space. This methodology, while analogous to traditional MCMC-based maximum likelihood training protocols for EBMs, deviates by substituting MCMC sampling with adversarial attacks for generating negative samples, thereby circumventing the challenges associated with gradient-based sampling in a discrete space. While there is no theoretical guarantee for the convergence of adversarial attacks to the model distribution, empirical evidence validates their efficacy in isolating spurious modes and generating statistically representative samples from the model distribution. By training the energy model to distinguish text samples from the target distribution from these perturbed samples, we enable it to suppress the spurious modes and more accurately allocate probability mass to the support of the target data distribution. Consequently, this allows us to produce higher-quality samples by first drawing from \( P_{LM} \) and subsequently applying adversarial perturbations to steer these base instances toward the support of the target distribution.

We assess the efficacy of our proposed method using an arithmetic sequence generation task, where it demonstrates a substantial performance enhancement compared to the baseline approach presented in \cite{deng2019residual}.

\section{Methodology}

\subsection{Notations and Definitions}
Let \( \mathcal{X} \) represent the input space and \( \mathbf{x} \in \mathcal{X} \) denote a specific instance of a text sequence. Importantly, \( \mathbf{x} \) can be represented as a sequence of words \( (x_1, x_2, \ldots, x_T) \), where \( T \) is the length of the sequence. We define \( P_{LM}(\mathbf{x}) \) as the probability of \( \mathbf{x} \) under the Locally Normalized Language Model (LNLM). Further, \( P_\text{data}(\mathbf{x}) \) denotes the empirical distribution of the target data. The ultimate objective of an Energy-Based Model (EBM) is to learn an energy function \( E_\theta(\mathbf{x}) \), parameterized by \( \theta \), which aims to assign lower energy values to the support of \( P_\text{data} \) and higher energy values to regions outside the support.

In the context of EBMs, probability distributions are formulated by linking the energy function \( E_\theta(x) \) to probabilities via a Gibbs distribution:
\begin{equation}
p_\theta(x) = \frac{\exp \left(-E_\theta(x)\right)}{Z(\theta)},
\end{equation}
where the normalizing constant \( Z(\theta) \), also known as the partition function, is an integral over the unnormalized probabilities across all states:
\begin{equation}
Z(\theta) = \int \exp \left(-E_\theta(x)\right) \, dx.
\end{equation}
In many complex models, the partition function \( Z(\theta) \) is computationally intractable, making the direct application of maximum likelihood estimation (MLE) for the model parameters \( \theta \) impractical. Traditional MLE for EBMs relies on computing the gradient of the log-likelihood function. Let \( p_{\text{data}} \) represent the distribution of the observed data; then the gradient of the log-likelihood can be expressed as:
\begin{equation}
\nabla_\theta \mathbb{E}_{x \sim p_{\text{data}}}\left[\log p_\theta(x)\right] = \mathbb{E}_{x \sim p_{\text{data}}}\left[\nabla_\theta (-E_\theta(x))\right] - \mathbb{E}_{x \sim p_\theta(x)}\left[\nabla_\theta (-E_\theta(x))\right].
\end{equation}
Intuitively, maximizing the log-likelihood using this gradient serves to reduce the energy function \( E_\theta(x) \) on samples from \( p_{\text{data}} \) while elevating it on samples drawn from \( p_\theta \). Samples of \( p_\theta \) are also known as \emph{negative samples}, and are typically generated with Markov Chain Monte Carlo (MCMC) sampling. However, as mentioned earlier, the discrete nature of text complicates the direct use of MCMC methods for text generation.

\subsection{Generating Negative Samples With Adversarial Attacks}
\label{sec:attack}
Given the challenges associated with sampling from \( p_\theta \) in a discrete textual space, we propose to generate negative samples by first drawing samples from  \( P_{LM}\) and then performing adversarial attacks on these base samples.

Let \( x_i \) denote the \( i^{th} \) token in the text sequence \( \mathbf{x} = (x_1, x_2, \ldots, x_T) \), which is sampled from \( P_{LM} \). The corresponding token embedding for \( x_i \) is represented as \( \mathbf{e}_i \), where \( \mathbf{e}_i \) belongs to \( \mathbb{R}^d \) and \( d \) signifies the dimensionality of the embedding space. To perform an adversarial attack, we first identify the index of the token that most significantly influences the energy function, as given by:

\begin{equation}
  i^* = \arg\max_{i=1,...,T}\|\nabla_{\mathbf{e}_i} (-E_\theta(x))\|_1
\end{equation}

Subsequently, we form a set of candidate sequences, denoted by \( \mathcal{C} \), by replacing \( x_{i^*} \) with tokens from its synonym set \( \mathbb{L}(x_{i^*}) \) within the vocabulary \( \mathcal{V} \). Specifically, each candidate sequence \( \mathbf{x'} \) is crafted by substituting \( x_{i^*} \) in the original sequence \( \mathbf{x} \) with an alternative token \( v \) from \( \mathbb{L}(x_{i^*}) \). Mathematically, each candidate sequence \( \mathbf{x'} \) can be expressed as:

\begin{equation}
  \mathbf{x'} = (x_1, ..., x_{i^*-1}, v, x_{i^*+1}, ..., x_T)
\end{equation}

The adversarial example is then selected as the candidate sequence that maximizes the negative energy function:

\begin{equation}
  \mathbf{x'}^* = \arg\max_{\mathbf{x'} \in \mathcal{C}} (-E_\theta(\mathbf{x'}))
\end{equation}

In line with the iterative nature of MCMC methods used in EBM training, we propose an iterative multi-step adversarial attack. Starting with an initial text sequence, each iteration identifies the most influential token based on the energy function's gradient. A new adversarial sample is crafted by replacing this token with a synonym to maximize the negative energy function. This new sample serves as the starting point for the subsequent iteration. The process continues iteratively, chaining together multiple attacks to incrementally exploit the model's spurious modes.


It should be noted that the above algorithm  is specifically tailored for the arithmetic sequence generation task, as outlined in \cref{sec:task}, and shares notable similarities with substitution-based text attacks prevalent in existing literature.
For real-world language modeling tasks, it may be advantageous to employ more sophisticated attacks, such as those outlined in \cite{morris2020textattack}

\subsection{Training the Energy Function \(E_\theta\)}

We outline the training algorithm employed to optimize the energy function \(E_\theta\) in \cref{alg:training-ebm}. The primary goal is to enable \(E_\theta\) to approximate the target distribution \( p_{\text{data}}\) by training the model to discriminate between samples drawn from \( p_{\text{data}} \) and negative samples generated with the attack algorithm as outlined in \cref{sec:attack}. 

  \begin{algorithm}
    \caption{Training Sequence Energy-Based Model with Iterative Multi-Step Adversarial Examples}
    \label{alg:training-ebm}
    \begin{algorithmic}[1]
    \STATE \textbf{Input:} Number of epochs \( N \), batch size \( B \), learning rate \( \alpha \), number of attack steps \( S \)
    \STATE \textbf{Initialize:} EBM parameters \( \theta \)
    
    \FOR{epoch \( = 1, 2, \ldots, N \)}
        \STATE Sample mini-batch \( \mathbf{X}_{\text{data}} \) from \( p_{\text{data}} \)
        \STATE Sample mini-batch \( \mathbf{X}_{\text{LM}} \) from \( P_{\text{LM}} \)
            
        \STATE \textit{\# Generate multi-step adversarial examples for each sample in mini-batch}
        \FOR{each \( \mathbf{x}_{\text{LM}} \) in \( \mathbf{X}_{\text{LM}} \)}
            \FOR{step \( = 1, 2, \ldots, S \)}
                \STATE Compute \( i^* = \arg\max_{i=1,...,T} \|\nabla_{\mathbf{e}_i} (-E_\theta(\mathbf{x}_{\text{LM}}))\|_1 \)
                \STATE Generate candidate sequences \( \mathcal{C} \) by replacing \( x_{i^*} \) with synonyms
                \STATE Compute \( \mathbf{x'}^* = \arg\max_{\mathbf{x'} \in \mathcal{C}} (-E_\theta(\mathbf{x'})) \)
                \STATE Set \( \mathbf{x}_{\text{LM}} \leftarrow \mathbf{x'}^* \) for the next step
            \ENDFOR
            \STATE Store the final \( \mathbf{x'}^* \) in set \( \mathbf{X'}^* \)
        \ENDFOR
            
        \STATE \textit{\# Update model parameters using the entire mini-batch}
        \STATE Compute gradient \( \nabla_\theta \sum_{\mathbf{x}_{\text{data}} \in \mathbf{X}_{\text{data}}} E_\theta(\mathbf{x}_{\text{data}}) \) and \( \nabla_\theta \sum_{\mathbf{x'}^* \in \mathbf{X'}^*} E_\theta(\mathbf{x'}^*) \)
        \STATE Update \( \theta \) using gradient descent:
        \[
        \theta \leftarrow \theta - \alpha \left( \nabla_\theta \sum_{\mathbf{x}_{\text{data}} \in \mathbf{X}_{\text{data}}} E_\theta(\mathbf{x}_{\text{data}}) - \nabla_\theta \sum_{\mathbf{x'}^* \in \mathbf{X'}^*} E_\theta(\mathbf{x'}^*) \right)
        \]
    \ENDFOR
    \end{algorithmic}
  \end{algorithm}
  
\subsection{Improving Generation With the Energy Function \(E_\theta\)}

After training both \(P_{LM}\) and \(E_\theta\), we enhance the quality of generated samples by initially drawing from \(P_{LM}\) and subsequently applying adversarial attacks to these samples via the iterative attack algorithm described in Section \(\ref{sec:attack}\).  Intuitively, if \(E_\theta\) is a faithful approximation of the target distribution, it will assign lower energy values specifically to the support of the target distribution. Therefore, when the iterative adversarial attacks are employed to guide the perturbed samples toward these low-energy regions, the samples are expected to align more closely with the target distribution, thereby elevating their overall quality.


  


\newpage
\section{Experiments}
\subsection{Task}
\label{sec:task}
In this work, our primary aim is to empirically validate our proposed adversarial training approach using a synthetic arithmetic sequence generation task, identical to the sequence generation task investigated in \cite{ResidualEBM2023}. The arithmetic sequences in this dataset are constructed based on elementary addition operations, as demonstrated by the sequence format:
\[  \resizebox{\linewidth}{!}{%
\(a_{10}\ a_9\ a_8\ a_7\ a_6\ a_5\ a_4\ a_3\ a_2\ a_1 + b_{10}\ b_9\ b_8\ b_7\ b_6\ b_5\ b_4\ b_3\ b_2\ b_1 = c_{11}\ c_{10}\ c_9\ c_8\ c_7\ c_6\ c_5\ c_4\ c_3\ c_2\ c_1\)
}
\]
Here, \(a_i, b_i\), and \(c_i\) denote individual digits ranging between 0 and 9. 
The \(c\) sequence is a result of the element-wise addition of \(a\) and \(b\), following conventional rules for arithmetic addition, including carrying over, if needed:
\[
c_i = a_i + b_i \quad \text{for} \quad i=1,\ldots, 10
\]
\[
c_{11} = \text{carry-over from the last addition, defaults to 0}
\]
An example sequence looks like:
\[
4\ 6\ 8\ 0\ 1\ 5\ 6\ 8\ 2\ 3 + 2\ 3\ 9\ 0\ 9\ 2\ 2\ 6\ 4\ 5 = 0\ 7\ 3\ 3\ 2\ 2\ 6\ 5\ 1\ 6\ 1
\]
Here, the sequence \(a\) is \(4\ 6\ 8\ 0\ 1\ 5\ 6\ 8\ 2\ 3\) and sequence \(b\) is \(2\ 3\ 9\ 0\ 9\ 2\ 2\ 6\ 4\ 5\). We note that the language model \(P_{LM}\) is trained to generate the whole sequence \(a+b=c\), as opposed to only \(c\).

By employing this synthetic task, we can precisely assess the quality of generated sequences through their compliance with the arithmetic rules of addition.
To quantify performance, we employ an average correctness metric computed over multiple generated sequences. The metric is particularly tailored to the characteristics of the arithmetic sequence generation task. The correctness for a single generated sequence is calculated as:
\[
\text{correctness} = \frac{\text{number of correctly predicted digits in } c}{\text{total number of digits in } c}
\]
Here, a digit in sequence \( c \) is considered ``correctly predicted'' if it exactly coincides with the corresponding digit in the ground truth sequence.
Subsequently, the average correctness over \( N \) sequences is given by:
\[
\text{Average correctness} = \frac{1}{N} \sum_{i=1}^{N} \text{correctness}_i
\]
where \( \text{correctness}_i \) is the correctness of the \( i^{th} \) sequence. Thus, a perfect model would attain an average correctness of 1, indicating flawless sequence generation across all test instances.

By using this synthetic dataset, we aim to empirically demonstrate the efficacy of our proposed adversarial training approach in mitigating spurious modes, thereby enhancing the capabilities of \(P_{LM}\) in sequence generation tasks.
\subsection{Training}

We set up our training protocol based on  \cite{ResidualEBM2023}.
The full training details for \(P_{LM}\) and \(E_\theta\) (pretraining, after attaching a language model head) can be found in \cref{tab:hyperparameters}.
In terms of the training objective, both models are trained by maximizing the conditional likelihood of the next token \( x_{i+1} \) given the sequence of previous tokens \( x_1, x_2, \ldots, x_i \):

\begin{equation}
  \label{eq:seq_mle}
\mathcal{L}(\theta) = \sum_{i=1}^{T} \log P(x_{i+1} \mid x_1, x_2, \ldots, x_i; \theta).
\end{equation}

Following pretraining, \(E_\theta\) undergoes a fine-tuning process aimed at approximating the target distribution \(P_\text{data}\). In the baseline approach of Residual-EBM \citep{deng2019residual},  \(E_\theta\) is fine-tuned using the conventional binary cross-entropy loss function. This fine-tuning process utilizes 1000 samples drawn from \(P_\text{data}\) and 1000 synthetic samples from \(P_{LM}\), spanning 5 epochs with early stopping. In contrast, our proposed methodology fine-tunes \(E_\theta\) according to \cref{alg:training-ebm}, employing the same number of \(P_\text{data}\) and \(P_{LM}\) samples but extending the training to 55 epochs, also with early stopping.  We use 5 attack steps in \cref{alg:training-ebm}.

Owing to the inherent stochasticity involved in the training procedure, we note fluctuations in model performance when utilizing Algorithm \ref{alg:training-ebm}. To account for this variability, we conduct the experiment ten times and report both individual and aggregate performance metrics.

\begin{table}[]
  \caption{Training setups for \(P_{LM}\) and \(E_\theta\) (pretraining)}
  \label{tab:hyperparameters}
  \centering
  \resizebox{0.6\textwidth}{!}{%
  \begin{tabular}{@{}lll@{}}
    \toprule
     & \(P_{\text{LM}}\)                                                                                                         & \(E_\theta\) (pretraining)                                                                                                              \\ \midrule
    Architecture     & \begin{tabular}[c]{@{}l@{}}One-layer LSTM\\ Embedding Size = 256 \\ Hidden Layer Size = 256\end{tabular}                  & \begin{tabular}[c]{@{}l@{}}Two-layer Transformer \\ Attention Heads = 4 \\ Embedding Size = 128 \\ Hidden Layer Size = 128\end{tabular} \\ \midrule
    Optimization     & \begin{tabular}[c]{@{}l@{}}Optimizer = AdamW \\ Learning Rate = 0.001 \\ Weight Decay = 0.05 \\ Epochs = 500\end{tabular} & \begin{tabular}[c]{@{}l@{}}Optimizer = AdamW \\ Learning Rate = \(6 \times 10^{-4}\) \\ Epochs = 500\end{tabular}                       \\ \midrule
    Training Samples & 1,000                                                                                                                     & 100,000                                                                                                                                 \\ \bottomrule
    \end{tabular}%
  }
  \end{table}

\subsection{Results}

\begin{figure}[h!]
  \centering
  \begin{subfigure}[b]{0.49\textwidth}
    \includegraphics[width=\textwidth]{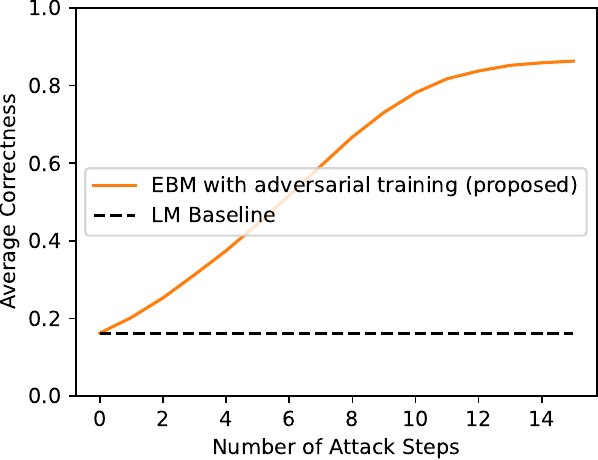}
    \caption{Average correctness of generated sequences as the number of attack steps varies}
    \label{fig:compare_a}
  \end{subfigure}
  \hfill 
  \begin{subfigure}[b]{0.49\textwidth}
    \includegraphics[width=\textwidth]{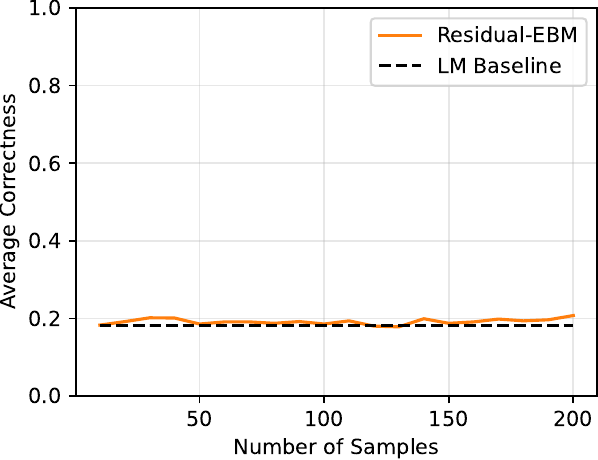}
    \caption{Average correctness of the generated sequence as the number of samples (in resampling) varies}
    \label{fig:compare_b}
  \end{subfigure}
  \caption{Performances of the proposed approach to learning EMBs with adversarial training and the baseline approach of Residual-EBM}
  \label{fig:compare}
\end{figure}

\cref{fig:compare} shows the performances of the proposed approach to learning EBMs with adversarial training and the baseline approach of Residual-EBM \citep{deng2019residual}. 
The Residual-EBM methodology employs importance sampling to produce novel sequence samples. To create a single sequence instance, the method initially draws multiple instances from \( P_{LM} \). Subsequently, it reweighs these instances using the learned energy function \( E_\theta \). Finally, it performs resampling based on the computed instance weights. \cref{fig:compare_b} illustrates the average correctness of generated sequences as a function of the number of instances sampled in this procedure. It can be seen that the importance sampling can only marginally improve the quality of generated sequences. In contrast, our approach significantly improves the quality of generated sequences as the number of attack steps increases.

\subsubsection{Ablation Study on Effect of Adversarial Training}
  In this ablation study we investigate the impact of adversarial training on the energy function. Specifically, we examine two distinct training paradigms:
  \begin{enumerate}
    \item \textbf{The Proposed Approach: EBM with Adversarial Training} ---
    In this setting, the energy function \(E_\theta\) is trained using samples from \(P_\text{data}\) as positive samples and \emph{adversarially perturbed} samples from \(P_{LM}\) as negative samples. Further details can be found in Algorithm \ref{alg:training-ebm}.
    \item \textbf{Baseline: EBM without Adversarial Training} --- 
    Here, the energy function \(E_\theta\) is trained using samples from \(P_\text{data}\) as positive samples and samples from \(P_{LM}\) as negative samples. Importantly, we do not apply adversarial perturbations to samples from \(P_{LM}\), making it functionally equivalent to setting the number of adversarial attack steps \(S\) to zero in \cref{alg:training-ebm}.
  \end{enumerate}
  After training \( E_\theta \), we generate test samples by first drawing from \( P_{LM} \) and then applying adversarial perturbations to the sampled data. The quality of the test samples is then evaluated using the average correctness metric as specified in Section~\ref{sec:task}.
  
\begin{figure}[H]
      \centering
      \includegraphics[width=0.5\linewidth]{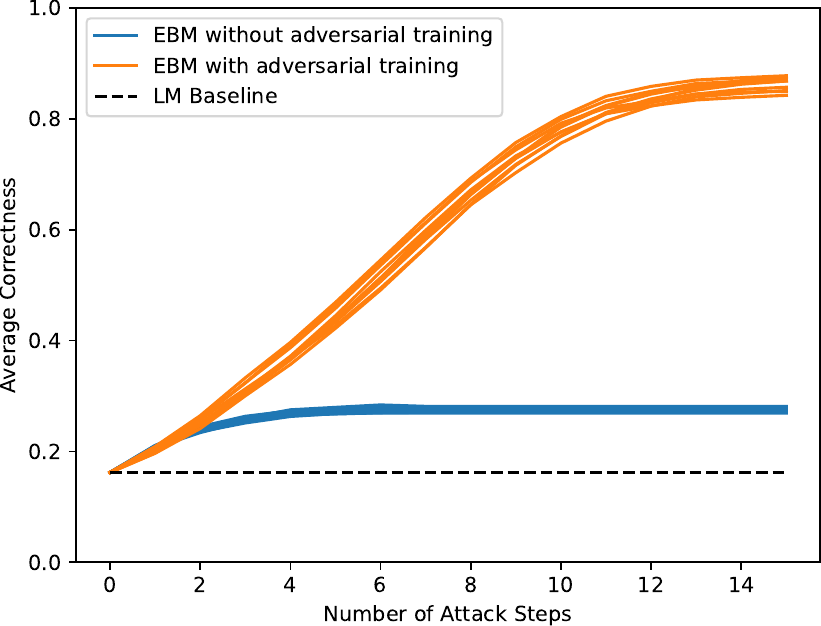}
      \caption{Average correctness of perturbed sequences as the number of  attack steps are varied}
      \label{fig:correct_curve}
  \end{figure}

  \cref{fig:correct_curve} illustrates the average correctness of the generated test samples as a function of the number of attack steps. In the absence of adversarial training, the quality of the generated samples experiences a marginal improvement after a few attack steps. In contrast, when attacking the EBM that has undergone adversarial training, the quality of the generated samples sees a substantial improvement.
  
\section{Conclusion}

This work demonstrates the potential for adversarial training techniques to enhance the training of energy-based models for discrete sequence generation tasks. By employing adversarial attacks to generate negative samples, the proposed methodology enables more effective suppression of spurious modes compared to prior binary classification-based training protocols.

However, the results also highlight key limitations and challenges associated with adversarial training that warrant further investigation. Most notably, the adversarial perturbation process introduces significant computational overhead compared to standard training paradigms. Each iteration of the attack algorithm requires evaluating the gradient of the energy function across all tokens in the sequence, followed by synonym substitutions and re-evaluation of the energy function. As the sequence length and size of the synonym set grow, these computations can become prohibitively expensive.
Developing efficient and scalable adversarial training protocols is an important area for future work towards enabling the application of discrete EBMs to complex, real-world NLG tasks.

In conclusion, this preliminary study demonstrates promising results but also highlights algorithmic and computational challenges associated with adversarial discrete EBM training that necessitate continued research and innovation. With further progress in this emerging field, adversarial techniques may prove instrumental in enhancing the generative capacities of discrete sequential models.

\bibliographystyle{abbrvnat}
\bibliography{references}

\end{document}